\documentclass[letterpaper, 10 pt, conference]{ieeeconf}  

\IEEEoverridecommandlockouts                              

\usepackage[noadjust]{cite}
\usepackage{amsmath,amssymb,amsfonts}
\usepackage{algorithmic}
\usepackage{graphicx}
\usepackage{textcomp}
\usepackage{xcolor}
\usepackage{caption}
\usepackage{multirow}
\usepackage{footnote}
\usepackage{hyperref}
\usepackage{array}
\usepackage{amssymb}
\usepackage{todonotes}
\usepackage[pscoord]{eso-pic}
\usepackage{flushend}

\title{\LARGE \bf
Towards agricultural autonomy: crop row detection under varying field conditions using deep learning
}

\author{Rajitha de Silva$^{1}$, Grzegorz Cielniak$^{2}$ and Junfeng Gao$^{3}$
\thanks{This work was supported by Lincoln Agri-Robotics as part of the Expanding Excellence in England (E3) Programme.}
\thanks{$^{1}$Rajitha de Silva, $^{2}$Grzegorz Cielniak and $^{3}$Junfeng Gao are with Lincoln Agri-Robotics Centre, Lincoln Institute for Agri-Food Technology, University of Lincoln, UK
        {\tt\small $^{1}$rajitha@ieee.org, $^{2}$gcielniak@lincoln.ac.uk, $^{3}$jugao@lincoln.ac.uk}}%
}

\begin{document}

\maketitle
\thispagestyle{empty}
\pagestyle{empty}

\begin{abstract}

This paper presents a novel metric to evaluate the robustness of deep learning based semantic segmentation approaches for crop row detection under different field conditions encountered by a field robot. A dataset with ten main categories encountered under various field conditions was used for testing. The effect on these conditions on the angular accuracy of crop row detection was compared. A deep convolutional encoder decoder network is implemented to predict crop row masks using RGB input images. The predicted mask is then sent to a post processing algorithm to extract the crop rows. The deep learning model was found to be robust against shadows and growth stages of the crop while the performance was reduced under direct sunlight, increasing weed density, tramlines and discontinuities in crop rows when evaluated with the novel metric.

\end{abstract}

\section{Introduction}
Computer vision algorithms has been identified as one of the key areas that need to be improved to promote the current agricultural systems\cite{oliveira2021advances}. Crop row detection is a key element in developing vision based navigation robots in agricultural robotics. Vision based crop row detection has been a popular research question in classical computer vision. Researchers have attempted to develop colour based segmentation of crops with various methods with varying levels of success\cite{romeo2012crop},\cite{guerrero2013automatic}. Recent work on crop row detection with deep learning based methods has been able to overcome the major challenges in implementing a real world vision based navigation system\cite{pang2020improved}, \cite{bah2019crownet}.

Individual work has been done to pursue various challenges in implementing a vision based navigation algorithms in crop rows. Researchers have attempted to address the effects of weed density, growth stages, shadows and discontinuities in crop row detection\cite{ji2011crop}, \cite{fue2020evaluation}. While these attempts indicate successful solutions to each of the problem, the need of a generic algorithm arises in practical implementations in such systems. A farmland may contain regions with varying levels of weed, various growth stages due to the soil conditions and different orientations of land. External environmental factors such as illuminations and shadows may also affect the performance of autonomous systems with the time of the day. It is therefore important to study the behaviour of a generic deep learning based crop row detection method under varying conditions. This understanding will assist the process of attaining fully autonomous farming in the future. 

 Unlike the usual semantic segmentation problems, crop row detection only require accurate prediction of a narrower crop line rather than accurate prediction of the vegetation region in an image. Therefore, traditional metrics in semantic segmentation are not perfectly suitable to reflect the capability of deep learning models to predict accurate crop rows. In this paper, we discuss the practical challenges encountered by a deep learning based crop row detection systems and the response of a generic deep learning based semantic segmentation model to such challenges. The need of a novel metric to ensure the capability of crop row detection will also be highlighted in this work. The annotation dataset with various field scenarios in this paper is publicly available in order to  further promote the development and optimisation of deep learning based row detection approaches for a reliable autonomous navigation system in fields. 

\section{Related Work}
Contour detection followed by Hough transform is a classic computer vision approach for crop row detection \cite{bonadies2019overview}. Hough transform based algorithms are popular among the researchers due to its accuracy of crop row detection \cite{winterhalter2018crop}, \cite{ji2011crop} and \cite{GAO201843}. However, these methods require fine tuning of threshold values to suit the varying ambient light conditions and growth stages. Thus, these classic approaches are not be suitable for real world vision-based navigation with varying environmental conditions. 

Ahmadi et al. \cite{ahmadi2020visual} has used excess green index \cite{woebbecke1995color} based algorithm to detect crop rows. Living tissue indicator and vegetation index has also been used to develop crop row detection algorithms\cite{bakker2008vision}, \cite{montalvo2012automatic}. These color based indicators are used to isolate pixels corresponding to plants in an image. However, these algorithms find it challenging to perform well in the presence of weeds due to their inability to separate crops and weeds. These methods are also incapable of extracting crop rows where crop canopy closes hiding background pixels corresponding to soil.

Recent developments in machine learning have enabled to perform crop row detection using neural networks, overcoming some of the barriers in classical computer vision. Lane detection in autonomous driving on roads are proven the capability of deep learning to be used in real life autonomous navigation \cite{zhao2020deep}. Adhikari et al. \cite{adhikari2020deep} has developed a enhanced skip network based autonomous navigation system for paddy fields. Their system is robust against shadows, field of view, row spacing and growth stage of the crop. However their system is implemented under the assumption that crop rows are clearly distinguishable in an image. Mean Pixel Deviation metric is used to evaluate the model. Their dataset is only limited to 350 images and it does not cover all the different scenarios that can occur in a field.

Bah et al. \cite{bah2019crownet} has used a fully convolutional network architecture that combines SegNet \cite{badrinarayanan2017segnet} and HoughCNet which is a Hough transform on a skeletonized binary image followed by a convolutional neural network (CNN). Their approach tries to eliminate the effects of crop row detection by weeds, discontinuities with multiple convolutional network stages. This complex architecture has been able to provide accurate crop row detection performance on images captured by unmanned aerial vehicles (UAV). In most of the deep learning based crop row detection approaches, mean Intersection over union (IoU) is considered as the key performance metric.

The existing work indicates the progress made in crop row detection over time and the advantages of deep learning based systems over classical computer vision approaches. The prior work on deep learning based crop row detection only analyze their model performance under one varying field parameter or without any variations, leading to the uncertainty of model deployment in field environments. The evaluation metrics used in previous work are also inherited from the general semantic segmentation evaluation metrics. The novel angle error metric presented in this paper proves the inability of classic metrics to reflect the performance of crop row detection with deep learning.

\section{Methodology}

The initial stage of the crop row detection system is based on semantic segmentation with deep learning. The predicted crop row mask is then used to estimate the crop rows in the image using Hough transform. The performance of the initial stage of this system is evaluated using a novel angle error metric.

The crop row detection algorithm is based on U-Net\cite{ronneberger2015u} architecture with a custom loss function to optimize angle error in crop row detection. U-Net has been one of the most popular image segmentation algorithm known for its ability to be trained with lesser amount of data and faster predictions. U-Net, being primarily intended for semantic segmentation, has been used to identify pixels belongs to a plant in a given image in agricultural applications\cite{fawakherji2019crop}. As apparent in Figure \ref{fig:lbl}, the expected outcome of using U-Net in this use case is to predict the line formed by the arrangement of plants in an input image. The crop row discontinuities due to missing plants are ignored and still assumed as a continuous crop line in the ground truth mask. 

\subsection{Training}
The U-Net was trained with a dataset of 1000 images which comprised of 100 images from each category listed in Table \ref{tab:cat}. The model was trained with binary cross entropy (BCE) loss with the Adam optimizer. The model was also trained separately with focal loss with $\gamma$ = 2. The predicted crop row masks were more sharper and narrower with the BCE loss. Therefore the BCE loss was selected for the work described in following sections.

\begin{figure}[t]
\centering
\captionsetup{justification=centering}
\includegraphics[scale=0.09]{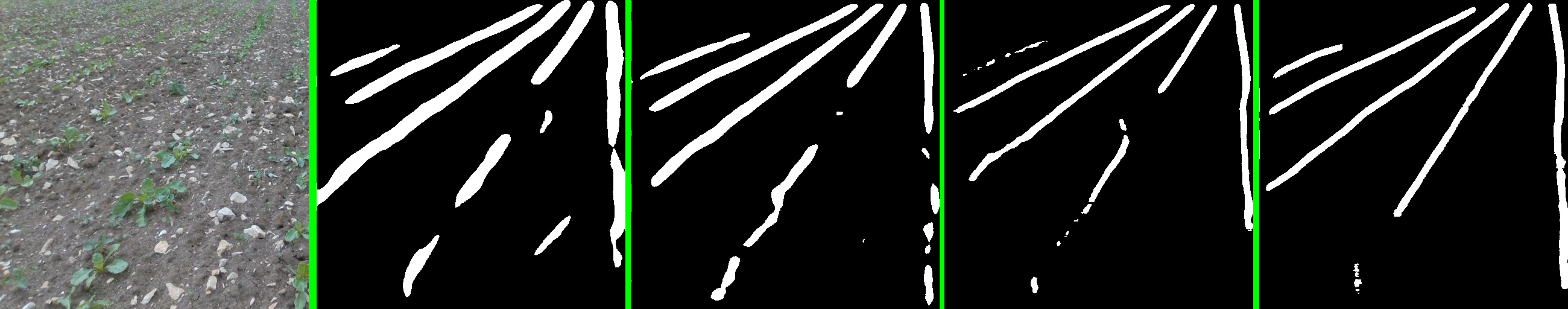}
\caption{Improvement of prediction at 5, 10, 20 and 40 epochs respectively (Left to Right)}
\label{fig:timp}
\end{figure}

According to Figure \ref{fig:timp}, U-Net has been trained to identify the crop rows with 5 epochs of training. However, this early stage model is only able to predict the line when there are no discontinuities in the crop row. It does not detect the crop row in regions where discontinuities are seen in the crop row. Its also noticeable that the predicted crop row is relatively wider than the ground truth mask. The model was able to eliminate false positives when training up to 10 epochs while it became more narrow and straight when trained up to 20 epochs. The model is able to predict the crop row despite the crop discontinuities at 40 epochs. The network was trained at 40 epochs where model could predict the crop rows by overcoming the aforementioned challenges.

\subsection{Angle Error Metric}

\begin{figure}[t]
\centering
\captionsetup{justification=centering}
\includegraphics[scale=0.3]{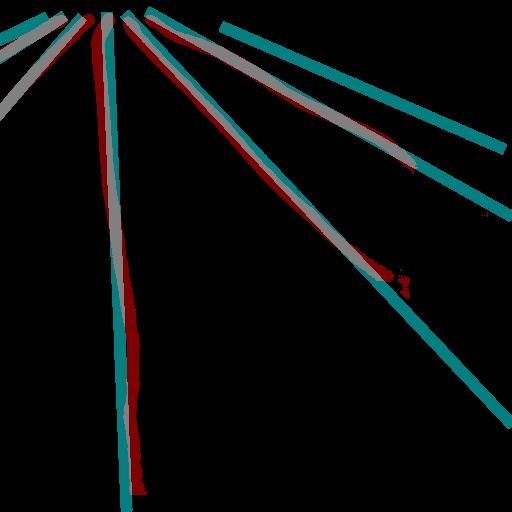}
\caption{IoU Visualization of the U-Net Prediction}
\label{fig:iou}
\end{figure}

Accuracy calculation in semantic segmentation involves pixelwise validation for black and white pixels in binary images. Therefore, accuracy metric mostly stay at higher values in crop row detection due to the data imbalance in ground truth masks. IoU is calculated with AND and OR operations over white pixels in a binary prediction mask and ground truth mask. IoU is often used as an alternative to accuracy metric which provide more reasonable value range to reflect the detection performance in image segmentation with class imbalance\cite{rahman2016optimizing}. Despite IoU being a more realistic metric to evaluate semantic segmentation models, it may not be the ideal metric for crop row detection. The IoU in crop row detection is dependent on the width of crop lines in the ground truth labels. Wider crop row labels would increase the IoU while narrower crop lines may decrease IoU. It may behave similarly with the width of the lines in predicted image. Figure \ref{fig:iou} shows the IoU visualization of ground truth image (blue) and the U-Net Prediction (red). The intersection region (gray) in the image still forms the relevant crop row mask despite the lower IoU value. Therefore, the angular offset between the predicted line and the ground truth would be a better metric for crop row detection based on semantic segmentation.

\begin{figure}[t]
\centering
\captionsetup{justification=centering}
\includegraphics[scale=0.75]{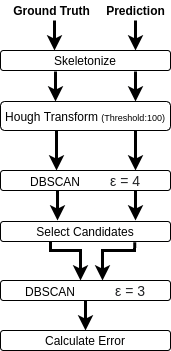}
\caption{Line angle error calculation}
\label{fig:cls}
\end{figure}

The angle error calculation is illustrated in Figure \ref{fig:cls}. The first stage of the angular error calculation is executed in two parallel and identical paths. The input image is skeletonized to be independent of the line width. Hough transform is used to determine the angles of each crop row in the image. The Hough transform threshold must be tuned to only predict one or a few lines for the images. The detected angles are clustered with DBSCAN\cite{schubert2017dbscan} to group Hough lines belonging to same crop row. The $\epsilon$ parameter in DBSCAN should be tuned to accommodate Hough lines belonging to same crop row within same category. One angle is selected as the representing candidate of the crop row where multiple Hough lines are detected per same crop row. The candidate is selection criteria was set in such a way that angle closest to the vertical direction is selected. Both the ground truth and prediction images are processed during the first stage of the angle error detection will generate two sets of angles corresponding to respective image. First stage of the proposed angle error calculation method was also used at post processing stage of the network to determine lines corresponding to crop rows from the segmented image. 

Angles detected from both the images are fed into another DBSCAN clustering stage. Each cluster generated by this stage will contain a pair of angles each belonging to the same crop row from ground truth image and prediction image. Angle error calculation is given in Equation \ref{eq:1} where k is the number of clusters identified by second DBSCAN stage with at least two values. $\theta$ represent the Hough line angles within each cluster.  This method is unable to calculate angle error with all the existing lines in an image. However, the missed lines will not affect the angle error calculation of individual image as the average of all the line errors are considered.

\begin{equation} \label{eq:1}
 Angle Error = \frac{1}{k}\sum_{n=1}^{k} | \theta_{n,max}-\theta_{n,min} |
\end{equation}

\section{Dataset}

\begin{figure}[t]
\centering
\captionsetup{justification=centering}
\includegraphics[scale=0.22]{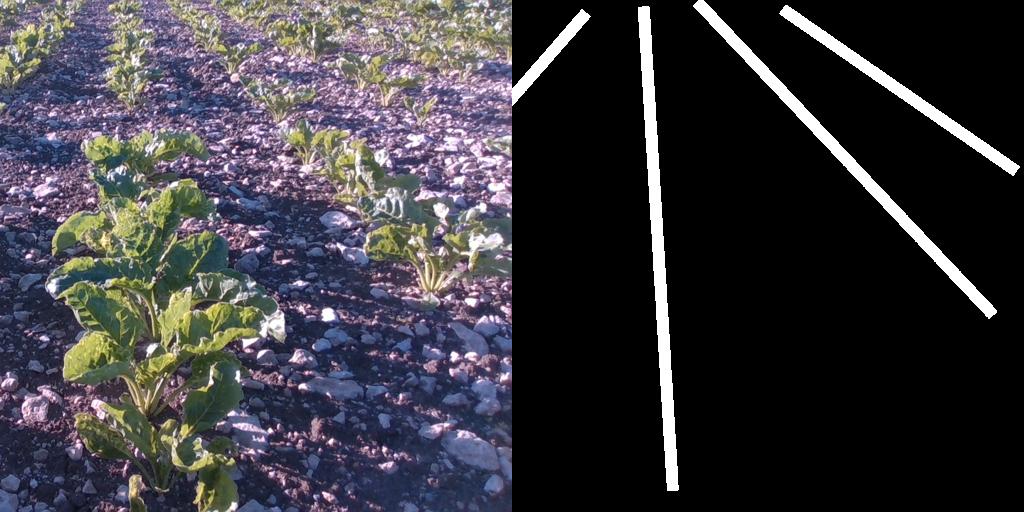}
\caption{Sample image and respective ground truth label mask}
\label{fig:lbl}
\end{figure}

The Crop Row Detection Lincoln Dataset (CRDLD) was created to gather a comprehensive training set for deep learning model which include multiple possible under varying field conditions. The dataset will include variations in weed density, shadows, sunlight, terrain elevation, growth stages, shape of the crop row and discontinuities in crop rows. CRDLD consists of 2000 images augmented from 500 base images. Each base image is cropped in different orientations to generate four images. The dataset is created by data recordings from 3 days within a span of two weeks under varying weather conditions and different times of the day in a sugar beet field. The sugar beet plants had 4-10 unfolded leaves throughout the duration of data capturing. The dataset was  split in half for training and testing. A larger test set was required since the model is evaluated in multiple categories of data. An example image and its corresponding ground truth image are shown in Figure \ref{fig:lbl}

\subsection{Hardware Setup}
\begin{figure}[t]
\centering
\captionsetup{justification=centering}
\includegraphics[scale=0.25]{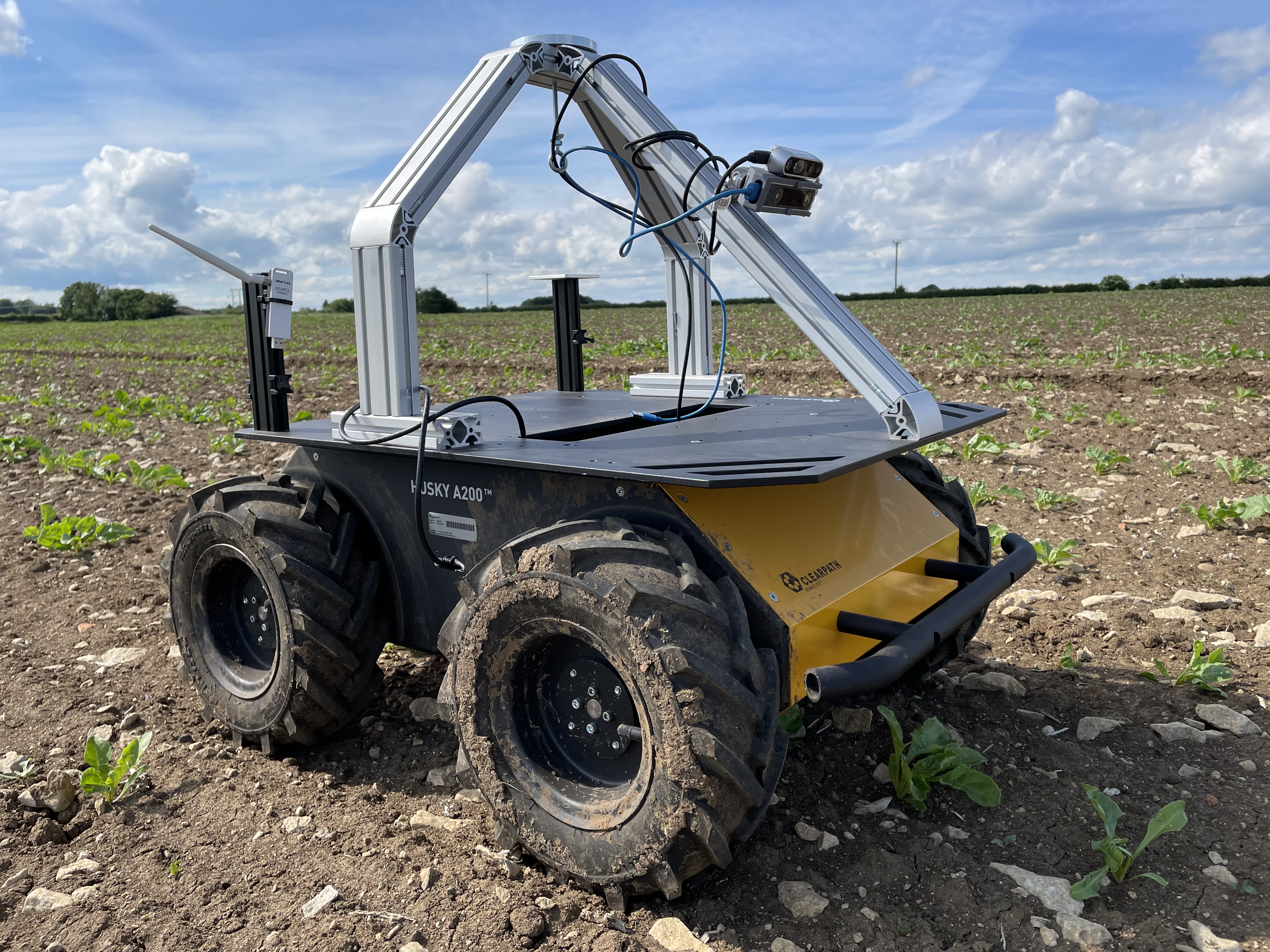}
\caption{Husky Robot with Realsense Cameras}
\label{fig:hus}
\end{figure}

The data collection was conducted using a Husky robot equipped with Intel RealSense D435i and T265 cameras as shown in Figure \ref{fig:hus}. RGB, infra red (IR) and depth images were captured with D435i camera. IR fish eye images and pose tracking data were captured with T265 camera. Only RGB images from D435i camera is used to generate the dataset in this paper.

\subsection{Data Categories}
\begin{figure*}[t]
\centering
\captionsetup{justification=centering}
\includegraphics[width=6in]{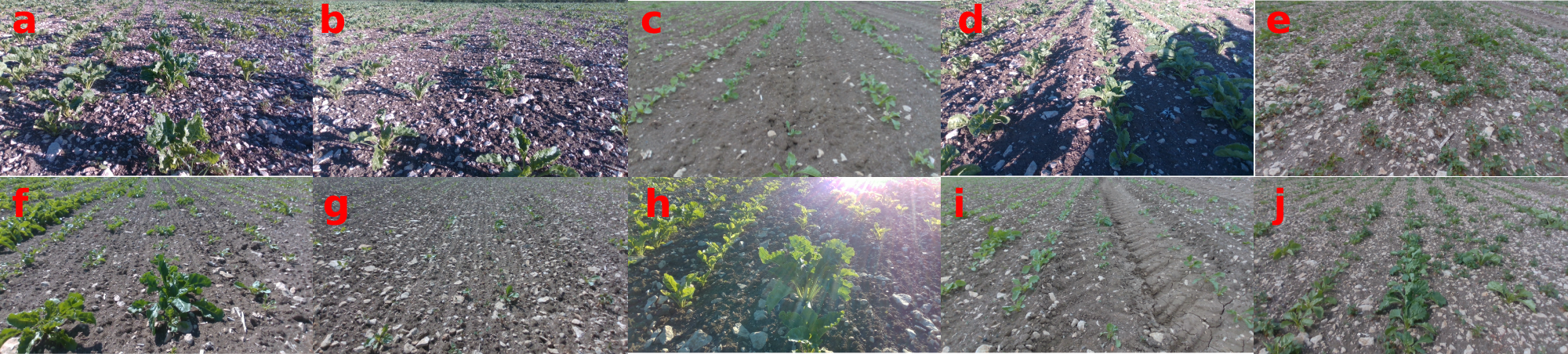}
\caption{Samples from data categories}
\label{fig:cat}
\end{figure*}

The RGB images in the dataset were classified into 10 categories depending on possible variations which could be expected in an open field farm due to varying weather, growth stages and time of the day. Each category contains 100 images along with the respective ground truth images\footnote{The dataset can be accessed with the following link: \textbf{\url{https://github.com/JunfengGaolab/CropRowDetection}}.}. The breakdown of categories is explained in Table \ref{tab:cat}. Sample images from each category are shown in Figure \ref{fig:cat}.

\begin{table}[t]
\centering
\caption{Data Categories}
\begin{center}
\begin{tabular}{|p{0.04\linewidth} | p{0.2\linewidth} | p{0.55\linewidth}|}
\hline
\textbf{ID} & \textbf{Name} & \textbf{Description}\\
\hline
\multirow{2}{0.4em}{a} & \multirow{2}{0.25em}{Horizontal Shadow} & Shadow falls perpendicular to the direction of the crop row \\ 
\hline
\multirow{3}{0.4em}{b} & \multirow{3}{0.25em}{Slope/ Curve} & Images captured while the crop row is not in a flat farmland or where crop rows are not straight lines \\ 
\hline
\multirow{2}{0.4em}{c} & \multirow{2}{0.25em}{Discontinuities} & Missing plants in the crop row which leads to discontinuities in crop row \\ 
\hline
\multirow{2}{0.4em}{d} & \multirow{2}{0.25em}{Front Shadow} & Shadow of the robot falling on the image captured by the camera \\ 
\hline
e & Dense Weed & Weed grown densely among the crop rows \\ 
\hline
\multirow{2}{0.4em}{f} & \multirow{2}{0.25em}{Large Crops} & Presence of one or many largely grown crops within the crop row \\ 
\hline
g & Small Crops & Crop rows at early growth stages \\ 
\hline
\multirow{2}{0.4em}{h} & \multirow{2}{0.25em}{Sunlight} & Sunlight falling on the camera causing lens flares and similar distortions \\ 
\hline
\multirow{2}{0.4em}{i} & \multirow{2}{0.25em}{Tyre Tracks} & Tyre tracks from tramlines running through the field \\ 
\hline
\multirow{2}{0.4em}{j} & \multirow{2}{0.25em}{Sparse Weed} & Sparsely grown weed scattered between the crop rows \\ 
\hline
\end{tabular}
\label{tab:cat}
\end{center}
\end{table}

These 10 categories of data could be identified as general challenges that a robot will have to overcome to perform reliable vision based navigation in an outdoor field. There have been many attempts to solve these challenges in individual dimensions. Montalvo et al. \cite{montalvo2012automatic} has developed a system to detect crop rows in maize fields with high weed pressure, hence trying to solve the crop row detection problem in category "e" of the dataset. Sivakumar et al. \cite{sivakumar2021learned} has developed an under canopy navigation robot with a vision based system attempting to address the major variation of appearance between early and late growth stages. Their approach is an attempt to solve the crop row detection problem in the categories "f" and "g" of the dataset. A compendious solution for the crop row detection problem is yet to be explored, despite the availability of different solutions to provide good performance in individual contexts. To this end, availability of highly diverse training dataset is vital to achieve a vision based navigation implementation that can overcome the challenges posed in a real world environment. 

\section{Results and Discussion}

\begin{figure}[t]
\centering
\captionsetup{justification=centering}
\includegraphics[scale=0.4]{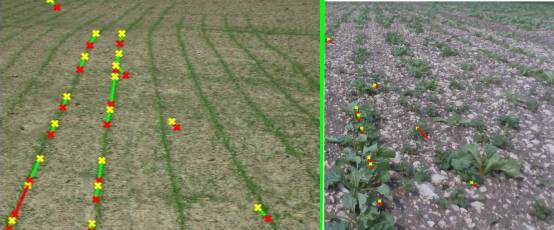}
\caption{Crop Row Detection Demonstration with Classic Computer Vision Methods\\ (Left image from CRBD Dataset\cite{vidovic2016crop})}
\label{fig:base}
\end{figure}

Figure \ref{fig:base} shows two predictions from Hough transform based crop row detection algorithm outlined in \cite{bonadies2019overview}. The algorithm uses contour detection to generate a crop row mask followed by Hough line transform. The algorithm failed to detect crop rows accurately when tested with images from the dataset (Figure \ref{fig:base} right image). However, their algorithm seemed to perform better when tested with the crop row images from Crop Row Benchmark Dataset (CRBD)\cite{vidovic2016crop}. It was observed that classic computer vision based crop row detection algorithms performed better with crop row images where crop row is narrow and compact such that the detected contours also become narrow leading to better detection accuracy with Hough line transform. However, such images could only obtained during early growth stages with lesser weed density.

The test dataset of CRDLD with 100 images per category was used to evaluate the model performance in different categories. The results listed in Table \ref{tab:pcat} indicate the performance of the model under each category. 

\begin{table}[t]
\centering
\caption{Model Performance in Data Categories}
\begin{center}
\begin{tabular}{|p{0.25\linewidth} | p{0.12\linewidth} | p{0.14\linewidth}| p{0.17\linewidth}|}
\hline
\textbf{Category Name} & \textbf{Accuracy} & \textbf{Mean IoU} & \textbf{Angle Error}\\
\hline
{Horizontal Shadow} & 89.70\% & 0.1990 & 0.0118\textdegree \\ 
\hline
{Slope/ Curve} & 88.01\% & 0.1905 & 0.0422\textdegree\\ 
\hline
{Discontinuities} & 88.49\% & 0.2672 & 0.0271\textdegree \\ 
\hline
{Front Shadow} & 89.88\% & 0.2708 & 0.0394\textdegree \\ 
\hline
Dense Weed & 90.01\% & 0.2056 & 0.0107\textdegree \\ 
\hline
{Large Crops} & 89.52\% & 0.2648 & 0.0170\textdegree \\ 
\hline
Small Crops & 88.97\% & 0.3127 & 0.0124\textdegree \\ 
\hline
{Sunlight} & 89.37\% & 0.1963 & 0.0087\textdegree \\ 
\hline
{Tyre Tracks} & 89.89\% & 0.2199 & 0.0207\textdegree \\ 
\hline
{Sparse Weed} & 89.72\% & 0.1980 & 0.0253\textdegree \\ 
\hline
\end{tabular}
\label{tab:pcat}
\end{center}
\end{table}

\begin{figure}[t]
\centering
\captionsetup{justification=centering}
\includegraphics[scale=0.22]{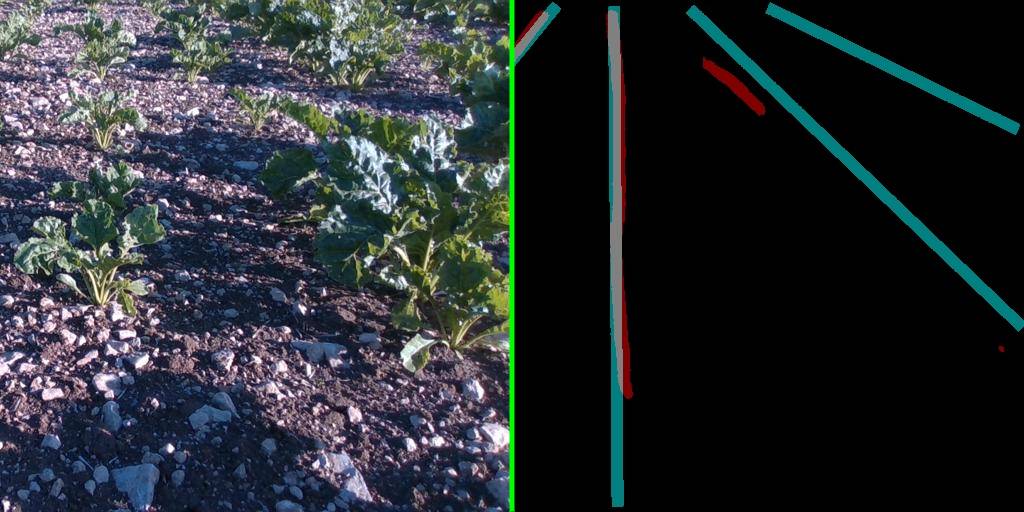}
\caption{Ground Truth vs. Prediction Comparison}
\label{fig:iouv}
\end{figure}

Mean accuracy of the network over all categories was 89.36\% while the mean IoU over all categories was recorded to be 0.2325. The lower IoU could be explained by the example in Figure \ref{fig:iouv}. The model tend to predict the crop rows in the middle of the image leaving out the crop row segments visible at the top corners of the image. However, these crop row segment may not be useful for navigation of a robot in a field. The predicted crop row and the ground truth might completely disconnect with a negligible offset in some instances, yet accurately predicting the angle of the relevant crop row. This contradiction between accuracy and IoU metrics indicate that their interpretation of the result is highly specific to the application scenario. Despite the contrariety in above metrics, overall angular error in predicted lines was 0.0215 degrees with none of the data categories exceeding an angular error of 0.05 degrees. 

\begin{figure}[t]
\centering
\captionsetup{justification=centering}
\includegraphics[scale=0.3]{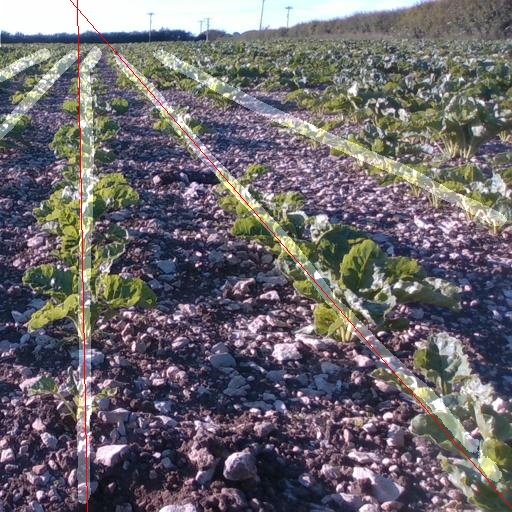}
\caption{Crop Row Detection Example in Slope/ Curve Category}
\label{fig:dcr}
\end{figure}

The worst crop row detection was observed in the Slope/ Curve category according to all the three metrics. Figure \ref{fig:dcr} shows the result from crop row detection in Slope/ Curve category. The white overlay mask is the prediction from U-Net model and the red color lines are generated by the post processing stage of the crop row detection algorithm.  It must be noted that the curved parts of the lines were predicted by the model. However, the post processing stage is only capable of detecting straight lines. 

\begin{figure}[t]
\centering
\captionsetup{justification=centering}
\includegraphics[scale=0.22]{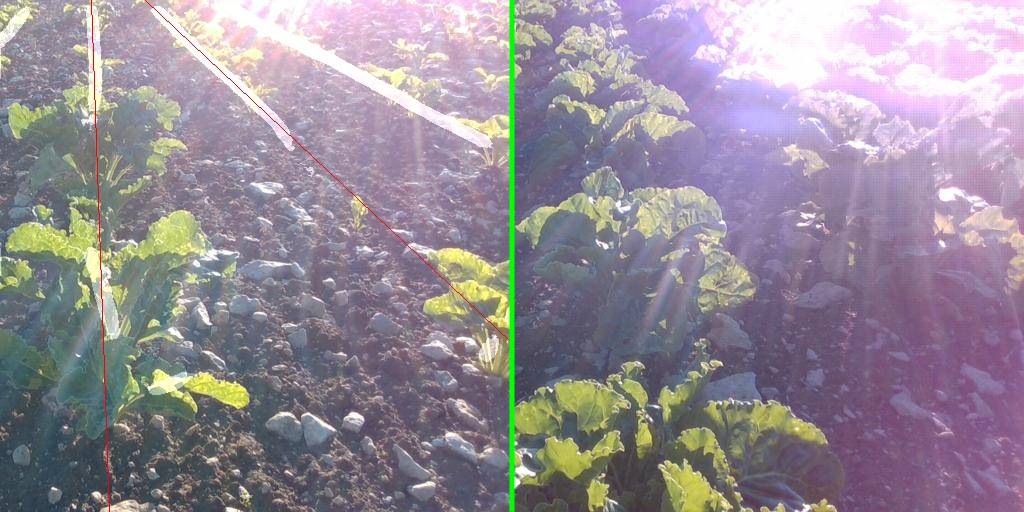}
\caption{Crop Row Detection Examples in Sunlight Category}
\label{fig:slr}
\end{figure}

Sunlight category contained examples which did not detect any crop rows, despite having the best performance according to angle error metric. Figure \ref{fig:slr} shows two examples with detected crop rows (left) and undetected crop rows (right). In images where sunlight was directly falling on to the camera were the hardest for the model to predict. Observation of cases where lens flare was present in the images proves that the effect of lens flares due to sunlight did not affect the performance of the model. The less angle error appeared in this category was due to the lack of detection of the lines.

\begin{figure}[t]
\centering
\captionsetup{justification=centering}
\includegraphics[scale=0.22]{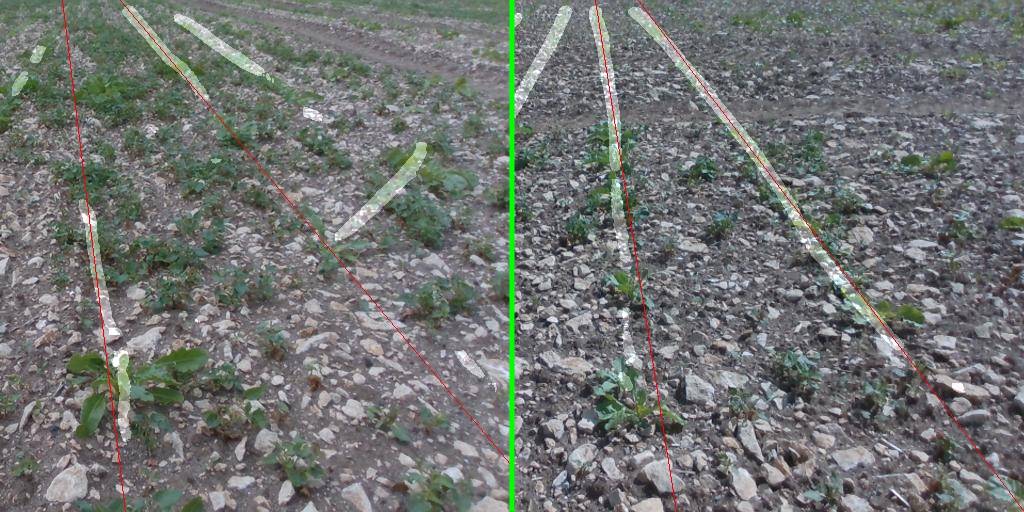}
\caption{Crop Row Detection Examples in Weed Categories}
\label{fig:wdr}
\end{figure}

Results from two categories dense weed and sparse weed are in Figure \ref{fig:wdr}. The left side example is from the dense weed category and the right example is from the sparse wed category. U-Net could not detect full lines in the dense weed category, yet it was able to predict the correct lines with a shorter length. The post processing algorithm was able to generate the crop rows with the shorted prediction masks. The right side example from sparse weed category depicts the ability of the U-Net model to separate weed from the crop rows.  

\begin{figure}[t]
\centering
\captionsetup{justification=centering}
\includegraphics[scale=0.3]{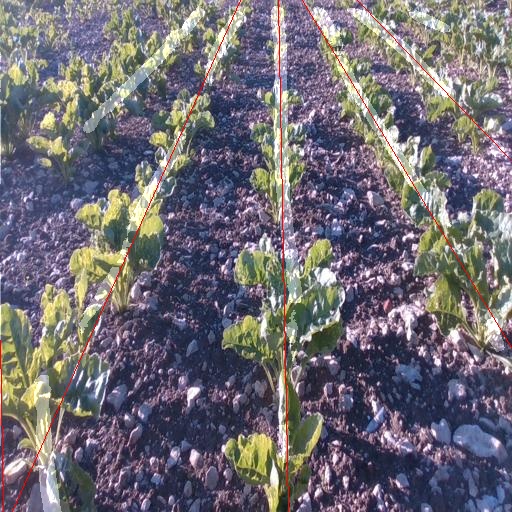}
\caption{Crop Row Detection Examples in Resized Global Image}
\label{fig:gbl}
\end{figure}

All the results discussed in this section were based on the 1000 augmented images from the set of 250 base images in the test dataset. Each prediction was made on a cropped image from the field of view (FOV) of the robot due to the input resolution of the U-Net model was limited to 512x512 pixels. The original images captured by the robot camera was in 16:9 aspect ratio. The original images were resized to 512x512 resolution to evaluate the model performance for the global image (with full FOV). All 250 images in the non-augmented test set were used to evaluate the model. The accuracy was recorded at 84.43\% with 0.1896 IoU and 0.0310 degrees angle error. The angle error increased with 0.0095 degrees when compared to the mean angle error of all categories in Table \ref{tab:pcat}. The accuracy and IoU metrics were also reduced by 4.93\% and 0.0429 respectively in comparison to the mean of all categories. This loss in model performance through all the metrics could be caused by the distortion introduced to the structure of the image during image resizing. Therefore predicting crop lines with separate cropped frames to stitch the global crop row mask would yield improved performance than the resized global images. A crop row detection example from the resized global images is shown in Figure \ref{fig:gbl}.

The results observed from the categories small crops and large crops were having relatively similar predictions. The growth stage of the crop did not have a significant impact of the prediction ability of the model. However, the images from earliest growth stages in the dataset had four or more unfolded leaves. Hence the variation in growth stages are in terms of size of the plant rather than having variation from very early growth stages. The model was only capable of filling the discontinuities of the crop rows during 18\% of the discontinuity category of the dataset. The behaviour of the predictions in tyre tracks category was also similar to the discontinuity category. The presence of shadows in the images did not have any significant effect on the predictions made by the U-Net model.

\begin{figure}[t]
\centering
\captionsetup{justification=centering}
\includegraphics[scale=0.15]{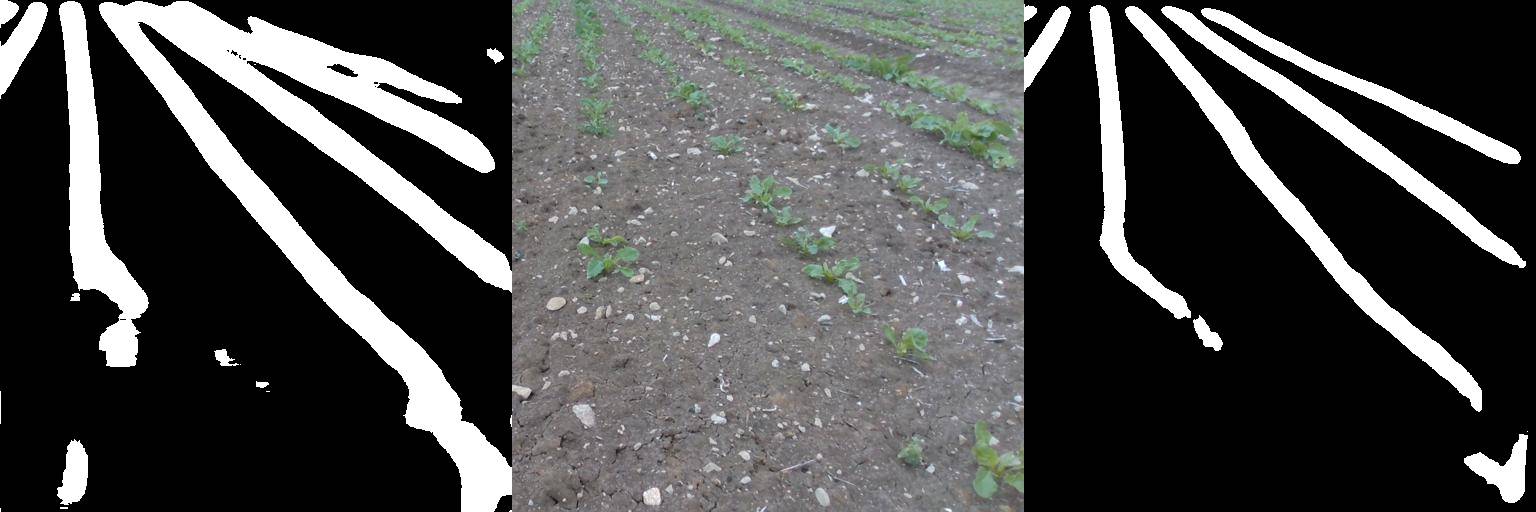}
\caption{Model Predictions from Focal loss (left) and BCE loss (right)}
\label{fig:fbce}
\end{figure}

A comparison of the predictions from the U-Net model with BCE loss and focal loss could be seen in Figure \ref{fig:fbce}. The mean accuracy, IoU and angle error for the model trained with focal loss are 89.72\%, 0.1285 and 0.0181 degrees respectively. The accuracy of both models are almost similar. The IoU of focal loss based model was reduced by 44.73\% while angle error was improved by only 15.81\%, in comparison to the BCE loss based model. Despite the slight improvement in angle error, BCE loss based model predictions were found to be more narrower and had less noise.

\section{Conclusion}

Deep learning based crop row detection under varying environmental and field conditions is discussed in this paper. A novel angular error metric was used to evaluate the U-Net model. The angle error metric evaluates the true ability of a semantic segmentation based crop row detection algorithm to predict crop rows accurately, in contrast to the classical metrics: accuracy and IoU. This paper also evaluates the capabilities of deep learning based crop row detection algorithm under various dimensions: shadows, weed density, discontinuities, sunlight and growth stages. It was determined that the model performance was independent of the shadows and growth stages. The model performance was slightly affected by the weed density and tyre tracks. The model was struggling in the event of discontinuities and direct sunlight in the camera. Semantic segmentation could be used to detect crop rows in real world applications with 0.05 degrees accuracy under varying field conditions. However, post processing algorithms for crop row extraction must be tuned depending on the field conditions.

The deep learning model used in this paper was implemented with its basic parameters without any modifications. The model could be improved with a better loss function which reflects the angle error in crop rows at later training stages. The performance of this method could further evaluated in different locations with different plant types for a broader understanding of its performance.
The post processing stage of the algorithm could be further improved with dynamic thresholds depending on the field conditions.

\bibliography{root}

\begin{thebibliography}{10}
\providecommand{\url}[1]{#1}
\csname url@samestyle\endcsname
\providecommand{\newblock}{\relax}
\providecommand{\bibinfo}[2]{#2}
\providecommand{\BIBentrySTDinterwordspacing}{\spaceskip=0pt\relax}
\providecommand{\BIBentryALTinterwordstretchfactor}{4}
\providecommand{\BIBentryALTinterwordspacing}{\spaceskip=\fontdimen2\font plus
\BIBentryALTinterwordstretchfactor\fontdimen3\font minus
  \fontdimen4\font\relax}
\providecommand{\BIBforeignlanguage}[2]{{%
\expandafter\ifx\csname l@#1\endcsname\relax
\typeout{** WARNING: IEEEtran.bst: No hyphenation pattern has been}%
\typeout{** loaded for the language `#1'. Using the pattern for}%
\typeout{** the default language instead.}%
\else
\language=\csname l@#1\endcsname
\fi
#2}}
\providecommand{\BIBdecl}{\relax}
\BIBdecl

\bibitem{oliveira2021advances}
L.~F. Oliveira, A.~P. Moreira, and M.~F. Silva, ``Advances in agriculture
  robotics: A state-of-the-art review and challenges ahead,'' \emph{Robotics},
  vol.~10, no.~2, p.~52, 2021.

\bibitem{romeo2012crop}
J.~Romeo, G.~Pajares, M.~Montalvo, J.~Guerrero, M.~Guijarro, and A.~Ribeiro,
  ``Crop row detection in maize fields inspired on the human visual
  perception,'' \emph{The Scientific World Journal}, vol. 2012, 2012.

\bibitem{guerrero2013automatic}
J.~M. Guerrero, M.~Guijarro, M.~Montalvo, J.~Romeo, L.~Emmi, A.~Ribeiro, and
  G.~Pajares, ``Automatic expert system based on images for accuracy crop row
  detection in maize fields,'' \emph{Expert Systems with Applications},
  vol.~40, no.~2, pp. 656--664, 2013.

\bibitem{pang2020improved}
Y.~Pang, Y.~Shi, S.~Gao, F.~Jiang, A.-N. Veeranampalayam-Sivakumar,
  L.~Thompson, J.~Luck, and C.~Liu, ``Improved crop row detection with deep
  neural network for early-season maize stand count in uav imagery,''
  \emph{Computers and Electronics in Agriculture}, vol. 178, p. 105766, 2020.

\bibitem{bah2019crownet}
M.~D. Bah, A.~Hafiane, and R.~Canals, ``Crownet: Deep network for crop row
  detection in uav images,'' \emph{IEEE Access}, vol.~8, pp. 5189--5200, 2019.

\bibitem{ji2011crop}
R.~Ji and L.~Qi, ``Crop-row detection algorithm based on random hough
  transformation,'' \emph{Mathematical and Computer Modelling}, vol.~54, no.
  3-4, pp. 1016--1020, 2011.

\bibitem{fue2020evaluation}
K.~Fue, W.~Porter, E.~Barnes, C.~Li, and G.~Rains, ``Evaluation of a stereo
  vision system for cotton row detection and boll location estimation in direct
  sunlight,'' \emph{Agronomy}, vol.~10, no.~8, p. 1137, 2020.

\bibitem{bonadies2019overview}
S.~Bonadies and S.~A. Gadsden, ``An overview of autonomous crop row navigation
  strategies for unmanned ground vehicles,'' \emph{Engineering in Agriculture,
  Environment and Food}, vol.~12, no.~1, pp. 24--31, 2019.

\bibitem{winterhalter2018crop}
W.~Winterhalter, F.~V. Fleckenstein, C.~Dornhege, and W.~Burgard, ``Crop row
  detection on tiny plants with the pattern hough transform,'' \emph{IEEE
  Robotics and Automation Letters}, vol.~3, no.~4, pp. 3394--3401, 2018.

\bibitem{GAO201843}
J.~Gao, W.~Liao, D.~Nuyttens, P.~Lootens, J.~Vangeyte, A.~Pi{\v{z}}urica,
  Y.~He, and J.~G. Pieters, ``Fusion of pixel and object-based features for
  weed mapping using unmanned aerial vehicle imagery,'' \emph{International
  journal of applied earth observation and geoinformation}, vol.~67, pp.
  43--53, 2018.

\bibitem{ahmadi2020visual}
A.~Ahmadi, L.~Nardi, N.~Chebrolu, and C.~Stachniss, ``Visual servoing-based
  navigation for monitoring row-crop fields,'' in \emph{2020 IEEE International
  Conference on Robotics and Automation (ICRA)}.\hskip 1em plus 0.5em minus
  0.4em\relax IEEE, 2020, pp. 4920--4926.

\bibitem{woebbecke1995color}
D.~M. Woebbecke, G.~E. Meyer, K.~Von~Bargen, and D.~A. Mortensen, ``Color
  indices for weed identification under various soil, residue, and lighting
  conditions,'' \emph{Transactions of the ASAE}, vol.~38, no.~1, pp. 259--269,
  1995.

\bibitem{bakker2008vision}
T.~Bakker, H.~Wouters, K.~Van~Asselt, J.~Bontsema, L.~Tang, J.~M{\"u}ller, and
  G.~van Straten, ``A vision based row detection system for sugar beet,''
  \emph{Computers and electronics in agriculture}, vol.~60, no.~1, pp. 87--95,
  2008.

\bibitem{montalvo2012automatic}
M.~Montalvo, G.~Pajares, J.~M. Guerrero, J.~Romeo, M.~Guijarro, A.~Ribeiro,
  J.~J. Ruz, and J.~Cruz, ``Automatic detection of crop rows in maize fields
  with high weeds pressure,'' \emph{Expert Systems with Applications}, vol.~39,
  no.~15, pp. 11\,889--11\,897, 2012.

\bibitem{zhao2020deep}
Z.~Zhao, Q.~Wang, and X.~Li, ``Deep reinforcement learning based lane detection
  and localization,'' \emph{Neurocomputing}, vol. 413, pp. 328--338, 2020.

\bibitem{adhikari2020deep}
S.~P. Adhikari, G.~Kim, and H.~Kim, ``Deep neural network-based system for
  autonomous navigation in paddy field,'' \emph{IEEE Access}, vol.~8, pp.
  71\,272--71\,278, 2020.

\bibitem{badrinarayanan2017segnet}
V.~Badrinarayanan, A.~Kendall, and R.~Cipolla, ``Segnet: A deep convolutional
  encoder-decoder architecture for image segmentation,'' \emph{IEEE
  transactions on pattern analysis and machine intelligence}, vol.~39, no.~12,
  pp. 2481--2495, 2017.

\bibitem{ronneberger2015u}
O.~Ronneberger, P.~Fischer, and T.~Brox, ``U-net: Convolutional networks for
  biomedical image segmentation,'' in \emph{International Conference on Medical
  image computing and computer-assisted intervention}.\hskip 1em plus 0.5em
  minus 0.4em\relax Springer, 2015, pp. 234--241.

\bibitem{fawakherji2019crop}
M.~Fawakherji, A.~Youssef, D.~Bloisi, A.~Pretto, and D.~Nardi, ``Crop and weeds
  classification for precision agriculture using context-independent pixel-wise
  segmentation,'' in \emph{2019 Third IEEE International Conference on Robotic
  Computing (IRC)}.\hskip 1em plus 0.5em minus 0.4em\relax IEEE, 2019, pp.
  146--152.

\bibitem{rahman2016optimizing}
M.~A. Rahman and Y.~Wang, ``Optimizing intersection-over-union in deep neural
  networks for image segmentation,'' in \emph{International symposium on visual
  computing}.\hskip 1em plus 0.5em minus 0.4em\relax Springer, 2016, pp.
  234--244.

\bibitem{schubert2017dbscan}
E.~Schubert, J.~Sander, M.~Ester, H.~P. Kriegel, and X.~Xu, ``Dbscan revisited,
  revisited: why and how you should (still) use dbscan,'' \emph{ACM
  Transactions on Database Systems (TODS)}, vol.~42, no.~3, pp. 1--21, 2017.

\bibitem{sivakumar2021learned}
A.~N. Sivakumar, S.~Modi, M.~V. Gasparino, C.~Ellis, A.~E.~B. Velasquez,
  G.~Chowdhary, and S.~Gupta, ``Learned visual navigation for under-canopy
  agricultural robots,'' \emph{arXiv preprint arXiv:2107.02792}, 2021.

\bibitem{vidovic2016crop}
I.~Vidovi{\'c}, R.~Cupec, and {\v{Z}}.~Hocenski, ``Crop row detection by global
  energy minimization,'' \emph{Pattern Recognition}, vol.~55, pp. 68--86, 2016.

\end{thebibliography}
\end{document}